\documentclass[lettersize,journal]{IEEEtran}
\usepackage{amsmath,amsfonts}
\usepackage{algorithmic}
\usepackage{algorithm}
\usepackage{array}
\usepackage{booktabs}
\usepackage[caption=false,font=normalsize,labelfont=sf,textfont=sf]{subfig}
\usepackage{textcomp}
\usepackage{stfloats}
\usepackage{url}
\usepackage{verbatim}
\usepackage{graphicx}
\usepackage{makecell}
\usepackage[export]{adjustbox}
\usepackage{cite}
\usepackage{multirow}
\usepackage{tikz}
\usepackage{multicol}
\usepackage[abs]{overpic}
\usepackage[hidelinks,colorlinks=false]{hyperref}
\newcounter{panel}[figure]

\usepackage{scalerel}
\usetikzlibrary{svg.path}

\definecolor{orcidlogocol}{HTML}{A6CE39}
\tikzset{
  orcidlogo/.pic={
    \fill[orcidlogocol] svg{M256,128c0,70.7-57.3,128-128,128C57.3,256,0,198.7,0,128C0,57.3,57.3,0,128,0C198.7,0,256,57.3,256,128z};
    \fill[white] svg{M86.3,186.2H70.9V79.1h15.4v48.4V186.2z}
                 svg{M108.9,79.1h41.6c39.6,0,57,28.3,57,53.6c0,27.5-21.5,53.6-56.8,53.6h-41.8V79.1z M124.3,172.4h24.5c34.9,0,42.9-26.5,42.9-39.7c0-21.5-13.7-39.7-43.7-39.7h-23.7V172.4z}
                 svg{M88.7,56.8c0,5.5-4.5,10.1-10.1,10.1c-5.6,0-10.1-4.6-10.1-10.1c0-5.6,4.5-10.1,10.1-10.1C84.2,46.7,88.7,51.3,88.7,56.8z};
  }
}

\newcommand\orcidicon[1]{\href{https://orcid.org/#1}{\mbox{\scalerel*{
\begin{tikzpicture}[yscale=-1,transform shape]
\pic{orcidlogo};
\end{tikzpicture}
}{|}}}}

\begin{document}

\title{Contact-Based Fringe Projection Profilometry for High-Resolution 3-D Surface Measurement of Reflective and Transparent Objects}

\author{Ingu Yeo$^{\textsuperscript{\orcidicon{0009-0007-6105-9101}}}$, Hyung-Gun Chi$^{\textsuperscript{\orcidicon{0000-0001-5454-3404}}}$, and Jae-Sang Hyun$^{\textsuperscript{\orcidicon{0000-0003-1711-8243}}}$

\thanks{Corresponding author: Hyung-Gun Chi, Jae-Sang Hyun}
\thanks{Ingu Yeo and Jae-Sang Hyun are with the Department of Mechanical Engineering and the Yonsei Institute for Embodied Intelligence, Yonsei University, Seoul 03722, South Korea.}}

\markboth{IEEE Transactions on Instrumentation and Measurement}%
{Yeo \MakeLowercase{\textit{et al.}}: Contact-Based Fringe Projection Profilometry for Reflective and Transparent Objects}



\maketitle

\begin{abstract}
This paper presents a contact-based 3-D surface measurement method based on a Digital Fringe Projection (DFP) system, belonging to the vision-based tactile sensing family pioneered by the commercially successful GelSight sensor. Such sensors have proven effective for robotic fingertip manipulation and contact sensing. However, because GelSight employs photometric stereo with RGB LEDs, it does not measure absolute depth directly but instead infers it by integrating estimated surface gradients, which can accumulate reconstruction errors; in addition, it becomes increasingly difficult to calibrate as the sensing area grows, and its depth accuracy is challenged on highly reflective or transparent objects. To overcome these drawbacks, we propose a fringe-projection-based contact measurement technique that performs triangulation-based 3-D reconstruction on a coated silicone contact surface, providing dense per-pixel surface geometry and full-field 3-D shape measurement over the contact region. By integrating high-accuracy digital fringe projection into the sensor, our approach simplifies calibration over larger areas and enhances depth precision for complex surfaces. Experimental results, including a direct comparison with a GelSight Mini sensor, a sphere-fitting accuracy evaluation, and an uncertainty analysis, confirm that the proposed method significantly improves the accuracy and stability of structured-light-based 3-D measurements, allowing reliable reconstruction of objects with diverse optical properties.

\end{abstract}


\section{Introduction}
\IEEEPARstart{T}{actile sensors} provide contact-event shape and material information that is difficult to capture using vision alone, thereby enabling human--robot interaction and, in manufacturing, quantitative dimensional metrology and defect inspection. They therefore constitute an important measurement modality. Accurate tactile sensing that reproducibly estimates quantitative three-dimensional (3-D) surface geometry has become increasingly important with the rise of autonomous robotics and portable metrology~\cite{li2017high}. Consequently, diverse tactile sensing approaches, including vision-based optical tactile sensors and skin-type tactile arrays, have been actively developed~\cite{yuan2017gelsight, ward2018tactip, lambeta2020digit, donlon2018gelslim, wettels2008biomimetic, schmitz2011methods}. Among them, photometric-stereo-based tactile sensing has been widely adopted because it can provide dense contact-surface information using compact optical hardware. In this context, this study develops a tactile sensor based on Digital Fringe Projection (DFP), a structured-light optical measurement approach that directly reconstructs surface geometry through triangulation.

DFP is one of the key technologies that provides high-resolution and high-accuracy 3-D shape information in applications such as optical metrology and 3-D scanning~\cite{zhang2006high}. A typical DFP system operates as follows. First, a digital projector casts fringe patterns onto the surface of the object to be reconstructed. The patterns are deformed according to the object’s geometry. A camera then captures the deformed fringe images, and the captured images are used to reconstruct the 3-D shape through a triangulation method based on the pinhole model. For 3-D shape reconstruction, the phase-shifting method has been widely used~\cite{zuo2018phase}. Phase-shifting offers high accuracy in depth estimation while enabling rapid projection of multiple patterns, which supports high-speed depth mapping. Furthermore, it is robust to noise and distortions, which is beneficial for reconstructing accurate 3-D models.

However, there exists an inherent limitation in DFP-based 3-D reconstruction. Most critically, the projected pattern images must be properly “adhered” to the target surface for reliable phase retrieval~\cite{geng2011structured}. Therefore, 3-D surface measurement becomes difficult for highly reflective objects, strongly curved surfaces, and transparent objects, where fringe patterns saturate, do not adhere well, or become distorted by transmission and internal reflections. Numerous studies have attempted to overcome this issue. For example, the methods in~\cite{zhang2009high, feng2014general, feng2018high} are designed to obtain accurate 3-D data even in challenging situations, such as surfaces with very high or very low reflectivity. These approaches generate high dynamic range (HDR) data by utilizing multi-exposure imaging and combining images captured at different exposures. However, exposure alignment problems and errors introduced during the fusion process can degrade overall accuracy and constrain real-time applications. The study in~\cite{liu20113d} addresses reflection problems by minimizing specular effects through careful optical configuration and by using optical filters and wavelength-specific illumination, but the performance can be sensitive to the surface reflectance properties and the operating illumination conditions. Gamma correction has also been adopted to mitigate saturation by controlling light intensity~\cite{guo2004gamma}, although it does not guarantee optimal performance across all conditions. Similarly, multi-frequency fringe patterns have been employed to compensate nonlinear phase errors and achieve high accuracy in certain environments~\cite{lei2015multi}, but they increase system and processing complexity. Multi-exposure map fusion has been proposed to enable precise 3-D reconstruction\cite{li2022exposure}, at the cost of increased computational burden and storage requirements. In~\cite{feng2024multi}, the exposure time is adjusted according to surface properties to obtain single-frame HDR images and to integrate multi-view data, alleviating saturation; nevertheless, the system configuration and processing time can become substantial. As summarized in~\cite{wang2023saturation, wei2024method, li2023fringe}, many existing efforts focus on pattern design, exposure optimization, and algorithmic correction, but they are still constrained by the physical limitations of reflective or transmissive surfaces.

To address these limitations, this study develops a new tactile 3-D surface measurement method using DFP. A related direction is optical tactile sensing based on photometric stereo. In~\cite{yuan2017gelsight}, a photometric-stereo-based 3-D scanning system was proposed by exploiting elastomer properties. When a soft gel-like elastomer contacts the target surface, the elastomer deforms according to the applied pressure. One side of the elastomer is transparent while the other side is coated for controlled imaging, enabling the capture of contact images without saturation on metallic surfaces. By illuminating the deformed elastomer with LEDs and imaging it, 3-D information can be computed based on the intensity variations caused by illumination direction and geometry. In practice, such approaches estimate per-pixel surface normals (or gradients) from intensity changes and then integrate them to recover the 3-D shape.

However, the photometric-stereo-based approach in~\cite{woodham1980photometric, yuan2017gelsight} is advantageous mainly for local, static contact-based measurements and tends to show limitations for objects with large curvature or complex geometry. This is because the reflected intensity varies strongly with incident angle, and fixed illumination may not provide sufficient directional cues for reliable normal estimation over complex surfaces. In addition, GelSight does not directly measure the absolute depth along the z-axis; instead, it infers depth indirectly by estimating surface gradients and integrating them, which can accumulate reconstruction errors and lead to reduced depth accuracy. Furthermore, the look-up table (LUT) used in GelSight is typically built from experimentally collected data rather than being derived from a physically grounded reflectance model, which can limit flexibility under complex optical conditions and under variations in real surface reflectance. In contrast, the proposed method uses hardness-tunable silicone and a gray reflective coating to ensure that fringe patterns are adhered with high contrast even for highly reflective or transmissive objects. Moreover, we employ N-step phase-shifting in both horizontal and vertical directions together with Gray coding to improve robustness. After estimating intrinsic and extrinsic parameters through standard camera–projector stereo calibration using dot/board targets, we apply local Gaussian filtering to the wrapped phase and perform Gray-code unwrapping, maintaining stability at edges and high-curvature regions and enabling accurate tactile sensing.

To simultaneously overcome the above limitations and the inherent constraints of conventional DFP, this study develops a tactile 3-D surface measurement pipeline based on a coated silicone interface. First, by introducing a coated silicone layer, the proposed system alleviates the failure of active 3-D reconstruction caused by total reflection on metallic objects. Second, by applying DFP on the coated silicone surface, we achieve high-resolution and high-accuracy 3-D surface measurement. For highly reflective or transparent objects, the object is brought into contact with the silicone, and fringe patterns are projected onto the opposite (coated) side to acquire 3-D information. This design demonstrates clear experimental advantages. For metallic objects such as a nut, a T-bolt, and a metallic ball, conventional DFP fails due to saturation and GelSight shows edge loss and depth instability, whereas the proposed method forms fringes across the entire contact interface and achieves complete reconstruction without missing regions. For transparent objects such as a LEGO block and a bear ornament, as well as acrylic specimens, the proposed method preserves fine geometric details and performs full-field reconstruction without pattern loss caused by internal transmission. Furthermore, the cross-sectional dimensions of standard acrylic rods closely match vernier caliper measurements, supporting quantitative depth accuracy. Using the pinhole-model-based triangulation, one 3-D point can be obtained per camera pixel, and the phase-based nature of the reconstruction supports high-precision 3-D measurement, making it suitable for high-resolution surface measurement. Considering the high resolution and high accuracy of DFP, 3-D surface information can be effectively obtained, and as suggested in~\cite{mannsfeld2010highly, wang2013user, park2016mos2, jiang2024electrical}, combining a structured-light system with elastomers can enable more accurate tactile sensors. Moreover, as shown in~\cite{andrussow2023minsight, kim2024extremely}, this technology has potential to be extended to diverse applications including tactile interfaces, robotic skins, and wearable sensors.

To position the proposed sensor against the state of the art, we provide direct quantitative comparisons with representative baselines, including a photometric-stereo tactile sensor (GelSight) and conventional object-surface DFP. 

The comparison is reported using metrology-relevant metrics, including (i) dimensional error on calibrated artifacts, (ii) completeness and missing-region rate on specular and transparent targets, and (iii) repeatability statistics over repeated trials.

A preliminary version of this work was presented in a conference paper~\cite{yeo2025high}, which introduced the basic concept of contact-based DFP using a coated silicone. The present paper substantially extends that work by adding (i) a direct quantitative comparison against a commercial GelSight Mini sensor, (ii) a sphere-fitting-based accuracy evaluation on calibrated spherical artifacts, and (iii) a metrological uncertainty analysis with a combined and expanded uncertainty budget for the repeated measurements.

\section{Method}
\subsection{Tactile sensor structural design}
\begin{figure}[t]
\centering
\includegraphics[width=2.8in,keepaspectratio]{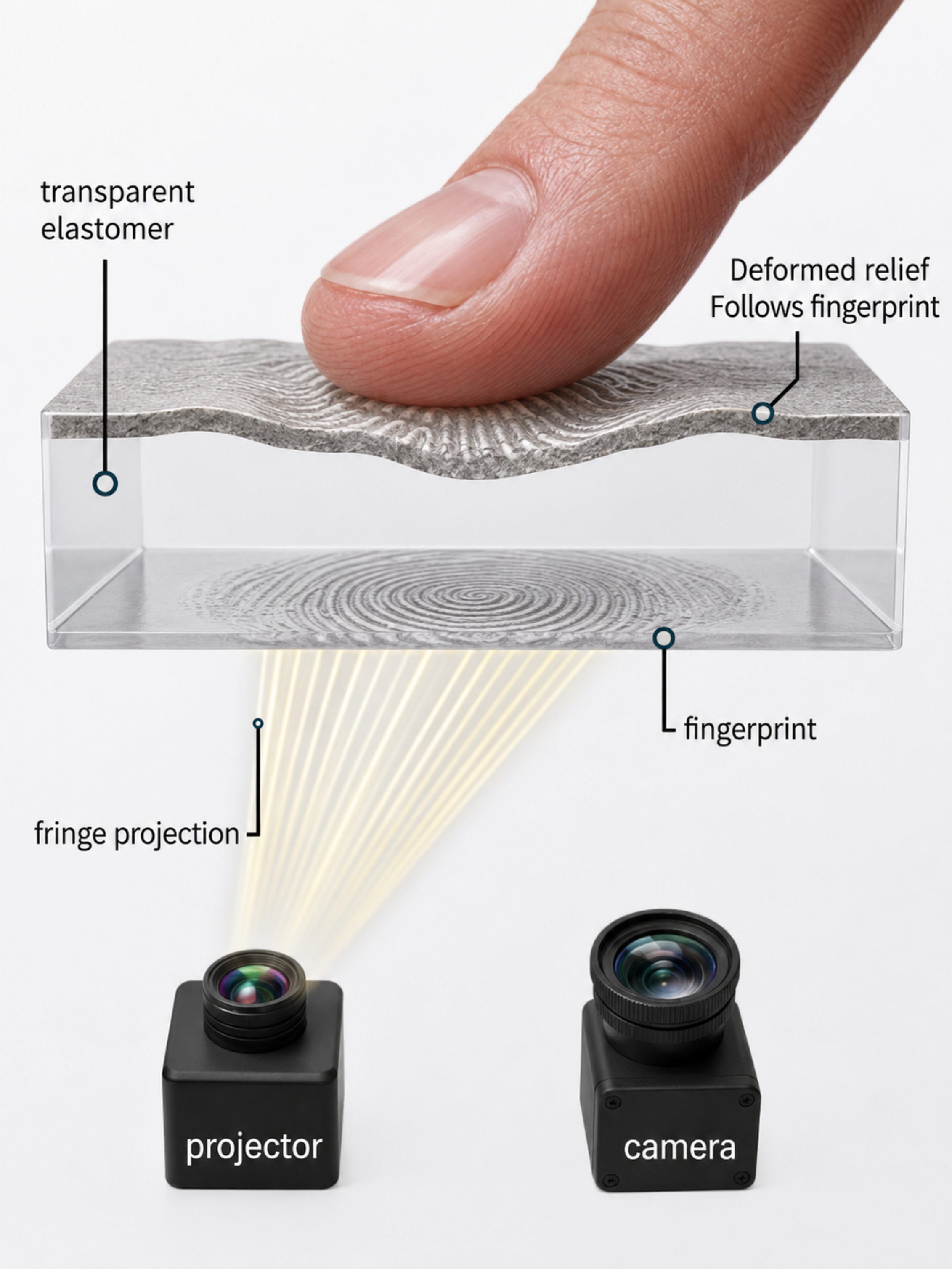}
\caption{Schematic of the proposed contact-based DFP sensing principle. The target object (here, a fingertip) contacts the transparent silicone, and the opposite gray-coated surface deforms to follow the contact relief. Fringe patterns are projected onto this coated surface and captured by the camera for triangulation-based 3-D reconstruction. (Conceptual schematic, not to scale.)}
\label{fig:concept}
\end{figure}

To conduct conventional DFP, a camera, a projector, a calibration board, and a target object are typically required for system calibration and 3-D reconstruction~\cite{zhang2000flexible}. To extend the applicability of DFP to metallic and transparent objects, we incorporate a silicone layer, inspired by vision-based tactile sensing approaches such as GelSight~\cite{yuan2017gelsight}. In such tactile sensors, the elastomer deforms according to the contacted surface geometry, allowing fine surface features to be indirectly captured through the deformation of the elastomer. In the proposed method, the silicone layer serves as a deformable optical interface: the target object contacts one side of the silicone, while fringe patterns are projected onto the coated opposite surface. As a result, objects on which fringe patterns cannot be directly formed can be measured through the deformation of the silicone surface, enabling high-resolution and high-accuracy 3-D surface reconstruction. Figure~\ref{fig:concept} schematically illustrates this operating principle.

1) \textit{Silicone hardness setting:} The base silicone slab was prepared using commercially available transparent liquid silicone. This silicone typically has a hardness of 20 when the base and curing agent are mixed in a $1:1$ ratio. However, the hardness can be adjusted by varying the ratio of the base to the curing agent. 

\begin{figure}
    \centering
    \includegraphics[width=3.1in, height=2.0in]{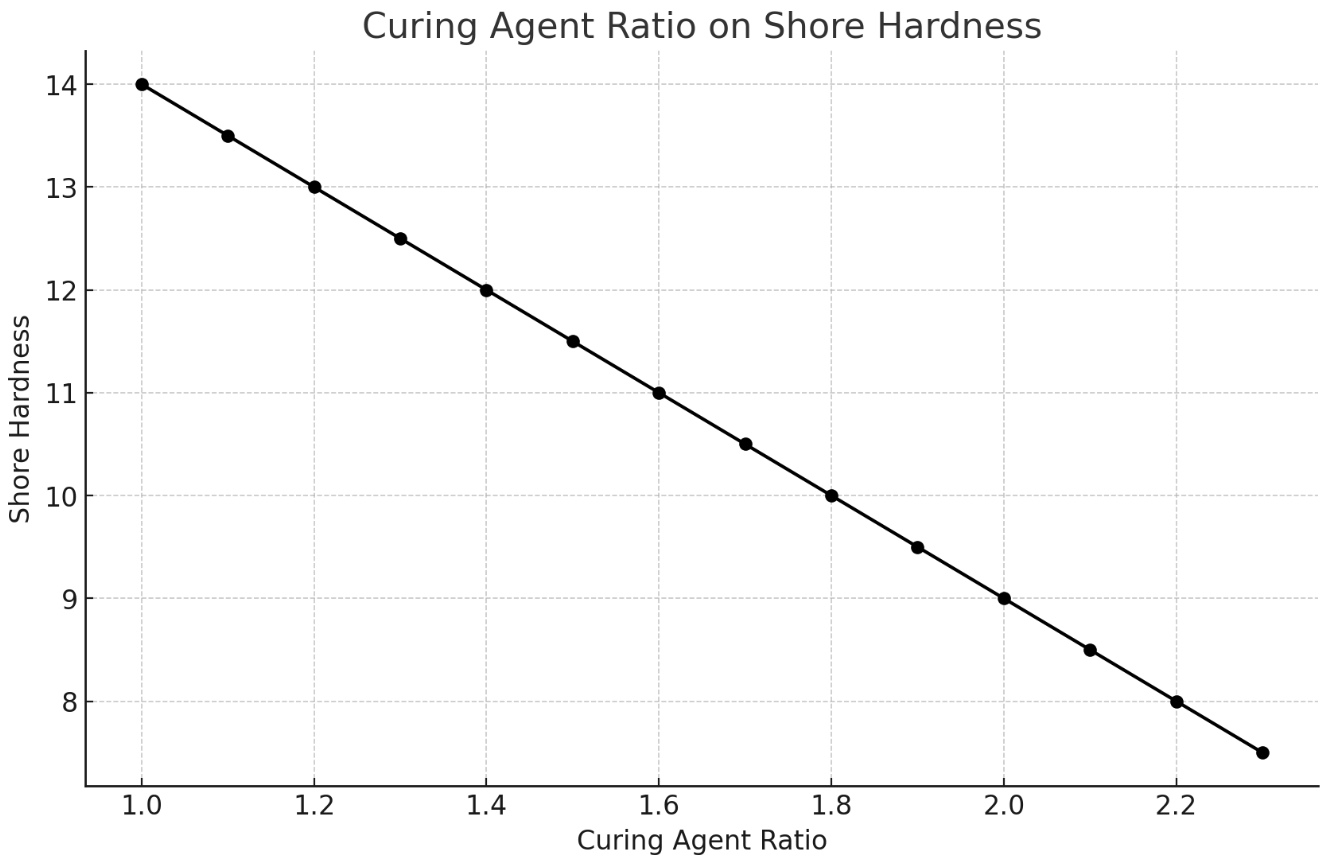}
    \caption{Silicone shore graph: Hardness to the mixing ratio of base and hardener}
    \label{graph}
\end{figure}

To identify the appropriate hardness for our experiments, we first optimized the base-to-curing-agent ratio. Three types of silicone were prepared: silicone A with a $1:1$ ratio, silicone B with a $1.5:1$ ratio, and silicone C with a $2:1$ ratio. The hardness of each specimen was measured with a Shore A durometer, and the resulting relationship between the mixing ratio and the measured hardness is summarized in Fig.~\ref{graph}.

Each silicone was then evaluated by contacting it with the surface of an object in our setup and observing its deformation along the sides to determine the most suitable hardness. To ensure consistency, all silicone samples were pressed under identical pressure using a Z-axis stage for uniform measurement. Under the same pressure, each silicone exhibited deformation according to the shape of the object. The measurement equipment used in the hardness characterization, dimensional evaluation, and comparative GelSight experiments is summarized in Table~\ref{tab:instrumentation}.

\begin{table}[!t]
\centering
\caption{Measurement equipment used in the experiments.}
\label{tab:instrumentation}
\footnotesize
\setlength{\tabcolsep}{2.5pt}
\renewcommand{\arraystretch}{1.1}
\begin{tabular}{p{0.18\columnwidth} p{0.26\columnwidth} p{0.42\columnwidth}}
\hline
Equipment & Model / Type & Role \\
\hline
Digital caliper & Mitutoyo CD-6''CSX & Reference width measurement of acrylic rods \\
Hardness tester & Shore A digital durometer & Hardness measurement of the silicone layer \\
Vision-based tactile sensor & GelSight Mini & Comparative reference sensor for side-by-side evaluation of 2-D/3-D tactile reconstruction performance \\
\hline
\end{tabular}
\end{table}

Among the tested samples, silicone C showed the best adaptability, conforming most effectively to the contact surface compared to silicone A and silicone B. 
Based on this evaluation, we selected the hardness of silicone C for the final setup. A square silicone sheet measuring 11 cm × 11 cm was fabricated and further cut into appropriately sized pieces to facilitate easy attachment and use in the experiment.






2) \textit{Selection of reflecting coating:} 
A gray reflective coating was prepared to be applied onto the silicone surface. The preparation procedure is as follows. First, a gray ink was formulated by appropriately mixing white and black pigments. Then, 0.2 g of catalyst, aluminum flakes, and 6 g of a silicone solvent were added to the mixture. A critical point in the preparation process is controlling the amount of aluminum flakes. If the amount is too small, fine surface features such as fingerprints may not be replicated when the object contacts the silicone. On the other hand, an excessive amount can lead to undesirable light reflection when projecting fringe patterns. Through multiple coating trials, an optimal quantity of aluminum flakes was determined. 

After thoroughly mixing all components, the coating was applied to one side of the silicone using an airbrush. This coating method, previously employed in~\cite{yuan2017gelsight,abad2019pilot,zhao2023gelsight}, plays a key role in enhancing optical performance by making the projected fringe patterns more clearly visible on the surface.



\subsection{3-D measurement method using DFP system}

1) \textit{Pinhole model for DFP system:}
Most optical devices like cameras and projectors are represented using the pinhole model. The pinhole model mathematically describes the relationship between 3-D world coordinates $(x, y, z)$ and 2D image coordinates, pixel $(u, v)$. The pinhole model can be expressed mathematically as follows:
\begin{equation}
s\begin{bmatrix}u\\v\\1\end{bmatrix}=\textbf{A}\left [ \textbf{R}| \textbf{t} \right ]\begin{bmatrix}x_{w}\\y_{w}\\z_{w}\\1\end{bmatrix}
\end{equation}
\begin{equation}
\textbf{A}=\begin{bmatrix}
f_{u}&\gamma & c_{u} \\
0&  f_{v}&  c_{v}\\
0&0 & 1\\
\end{bmatrix},\textbf{R}=\begin{bmatrix}r_{11} &r_{12}  &r_{13}  \\r_{21} &r_{22}  & r_{23}\\r_{31} & r_{32} &r_{33}  \\\end{bmatrix}, \textbf{t}=\begin{bmatrix}t_{1} \\t_{2}\\t_{3}\end{bmatrix}
\end{equation}

Here, $(u, v)$  represents the image pixel coordinates, and $(x, y, z)$ represents the 3-D world coordinates. $\textbf{A}$ is referred to as the intrinsic parameters, where $f_{u}$ and $f_{v}$ are the focal lengths, representing the distance from the lens center to the image sensor. $c_{u}$ and $c_{v}$ are the principal points, representing the intersection of the perpendiculars dropped from the pinhole onto the image sensor. $\gamma$ represents the degree to which the y-axis of the image sensor cell array is tilted. 

$\textbf{R}$ and $\textbf{t}$ are called extrinsic parameters, describing the transformation relationship between the camera coordinate system and the world coordinate system, expressed as rotation and translation transformations between the two coordinate systems. Unlike intrinsic parameters are not inherent to the device and may vary depending on how the camera and projectors are positioned and how the world coordinate system is defined.

When performing calibration with DFP, the intrinsic and extrinsic parameters of the camera and projector are determined. The projector is treated as an inverse camera, and the unwrapped phase is used to establish the correspondence between camera pixels and projector pixels. Based on this calibrated camera--projector correspondence, the 3-D world coordinates corresponding to each camera pixel $(u, v)$ are obtained through triangulation. Therefore, the phase values of the projected fringe patterns can be used to reconstruct the 3-D surface geometry.

2) \textit{Phase-shifting method:}
DFP has opted for a phase algorithm to locate the corresponding pixels between the camera and the projector. Although intensity-based methods are also common, the use of phase tends to be less susceptible to noise compared to intensity-based approaches. Additionally, it is less affected by surface reflection and texture surface color. 

To define the pixels of the output devices, the projector, as opposed to the input device, the camera, the horizontal and vertical fringe patterns must be projected onto the target and captured by the camera. The projected pattern should include at least three phase-shifted fringe patterns, known as the three-step phase-shifting algorithm \cite{zhang2018highbook}. However, to reduce measurement errors, we employ four or more shifted fringe images in our approach. When using N-step phase-shifted patterns, the fringe image can be represented as follows.

\begin{align} \label{eqn:my_equation}
    \begin{split}
    I_{1}(u,v) &= I'(u,v)+I''(u,v)\textrm{cos}(\phi (u,v)+\delta _{1})\\
    I_{2}(u,v) &= I'(u,v)+I''(u,v)\textrm{cos}(\phi (u,v)+\delta _{2})\\
    I_{3}(u,v) &= I'(u,v)+I''(u,v)\textrm{cos}(\phi (u,v)+\delta _{3})\\
    &\vdots \\ 
    I_{n}(u,v) &= I'(u,v)+I''(u,v)\textrm{cos}(\phi (u,v)+\delta _{n})
    \end{split}
\end{align}
To solve the equation and find the phase values, $I_{n}(u,v)$ represents the intensity of each fringe captured image at pixel coordinates \textit{(u,v)} of the camera. $I'_{n}(u,v)$ represents the mean intensity, and $I''_{n}(u,v)$ represents the intensity modulation at the pixel $(u,v)$, and $\delta$ is a shift for phase-shifting method algorithm. When using more than three fringe patterns, the least squares algorithm can be employed. Then, we can get the phase map, $\phi(u,v)$ is expressed as follows:
\begin{align}
\phi(u,v) = -\textrm{tan}^{-1}(\frac{\sum_{k=1}^{N}I_{k}\textrm{sin}(\delta_{k})}{\sum_{k=1}^{N}I_{k}\textrm{cos}(\delta_{k})})
\end{align}
However, the resulting phases from such searches exhibit discontinuities $-\pi$ at $\pi$. These discontinuities lead to multiple points with the same value, making it impossible to establish a one-to-one correspondence between pixels captured by the camera and those projected by the projector. To address the issue, a procedure called phase unwrapping is required, which renders the retrieved phase continuous and absolute. The phase unwrapping algorithm is expressed as follows,
\begin{equation}
    \Phi (u,v)= K(u,v)\times 2\pi+\phi (u,v)
\end{equation}
where $\phi(u,v)$ represents the wrapped phase, which contains discontinuities at the $-\pi$ to $\pi$ transition, and $K(u,v)$ represents the integer fringe order used to remove these discontinuities. The resulting continuous absolute phase is denoted by $\Phi(u,v)$.

3) \textit{Gray coding method:}
To obtain absolute phase, the process of phase unwrapping is required to unwrap the wrapped phase. There are several methods for phase unwrapping, and the method we employ is the Gray coding method \cite{zuo2016temporal}. The Gray coding phase-shifting technique designs unique codewords assigned to each $2\phi$ phase change period to determine the fringe order, i.e., $k(x,y)$. However, the maximum number of unique codewords is limited to $2^{M}$. To illustrate this more clearly, for example, three binary encoded patterns can generate up to eight unique codewords. By embedding codewords into the phase domain through phase shifting algorithms, more codewords can be generated. Specifically, the codewords and the unwrapping equation can be expressed as follows. Embed code word into phase with a stair phase function.
\begin{equation}
\phi _{s}=-\pi +Floor[x/T]\times2\pi/N
\end{equation}
Here $Floor[x/T]=k(x,y)$ generates the truncated integer representing fringe order that removes the decimals of a floating point data while keeps its integer part; $T$ the fringe pitch; the number of pixels per period; and N the total number of fringe periods.
Put the stair phase $\phi^{s}$ into three phase-shifted fringe patterns,

Obtain the stair phase from the coded fringe patterns through phase wrapping. Since the stair phase is encoded into the value ranging from $-\pi$ to $+\pi$, no spatial phase unwrapping is required
Determine fringe order $k(x,y)$ from the stair phase
\begin{equation}
    k(x,y) = \mathrm{Round}[N(\phi^{s}+\pi)/(2\pi)]
\end{equation}
Here Round[$x$] determines the closest integer.
Convert wrapped phase $\phi(x,y)$ to absolute phase. Once the fringe order $k(x,y)$ is determined, the wrapped phase can be unwrapped temporarily.

\section{Experiments}
To validate the proposed method, the following setup was established. The system consists of a CCD camera (FLIR BFS-U3-28S5M-C) equipped with a 50 mm focal length lens (Computar M5028-MPW3) and a Digital Light Processing (DLP) projector (Texas Instruments DLP3010EVM-LC). The camera resolution is set to $1936 \times 1464$, and the projector resolution is $1280 \times 720$. The setup is as shown in Fig.~\ref{fig:Setup}. The setup used aluminum extrusions as the main frame and was arranged in an up–down configuration to facilitate secure fixation of the object and the silicone. A z-axis stage was employed to attach the object to the silicone under a consistent normal load.

The first calibration procedure of the proposed method involves calibration of the camera and projector. The custom dot-patterned board for the designed system is used to calibrate the stereo camera and the projector. The projector projects vertical and horizontal fringe patterns onto the calibration board. The captured fringe patterns are then calibrated using a phase-shifting algorithm. 

The projected patterns consist of 18 sets of phase-shifted images and 7 Gray code images, each composed of vertical and horizontal patterns, along with an additional black-and-white pattern. The poses of 12 calibration boards are acquired. Furthermore, the experimental data presented here are filtered with an $8 \times 8$ Gaussian filter applied to the phase map to reduce the noise in the measurements.

We designed three experiments to evaluate whether the proposed method overcomes the inability to form fringe patterns on highly reflective and transparent materials and to verify quantitative accuracy by comparison with ground-truth dimensions, as well as to assess robustness relative to existing tactile sensors. First, we qualitatively validate performance using a metallic nut, a transparent LEGO block, and a metallic ball. Second, we quantitatively assess accuracy by comparing reconstructed cross-sections with the cross-sectional dimensions of standard acrylic rods. Third, we acquire and reconstruct data under identical scenarios with a GelSight Mini to contrast the impact on real outcomes.

\subsection{Metal objects and transparent objects}

\begin{figure}
    \centering
    \includegraphics[width=3.424in, keepaspectratio]{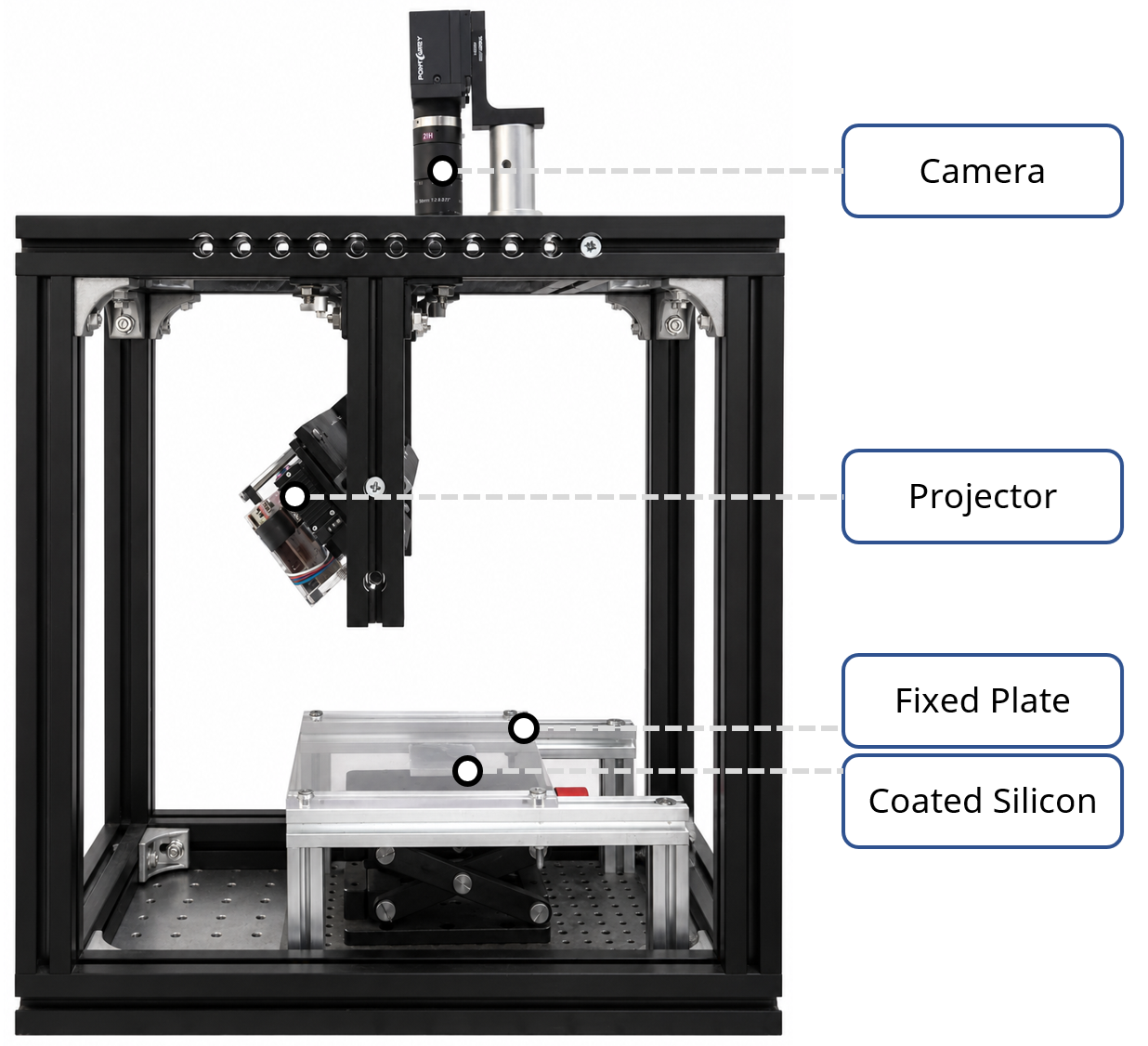}
    \caption{Photograph of our experimental system setup}
    \label{fig:Setup}
\end{figure}

\begin{figure*}[!t]
\centering
%
\refstepcounter{figure}
\setlength{\tabcolsep}{2pt}
\renewcommand{\arraystretch}{1.0}

\newcommand{\imgH}{1.3in}
\newcommand{\imgCellW}{0.145\linewidth}
\newcommand{\rowgap}{18pt}

\vspace{4pt} 

\newcommand{\rowlab}[2]{%
  \begin{minipage}[c][\imgH][c]{1.35cm}
    \centering
    \rotatebox{90}{\parbox{2.2cm}{\centering\textbf{#1}\\[-1pt]\scriptsize #2}}
  \end{minipage}%
}

\newcommand{\labimg}[2]{%
  \refstepcounter{panel}\label{#2}%
  \begin{minipage}[c]{\imgCellW}\centering
    \begin{tikzpicture}[baseline=(img.south)]
      \node[inner sep=0pt] (img) {\includegraphics[height=\imgH,keepaspectratio]{#1}};
      \node[anchor=north west, xshift=2pt, yshift=-2pt,
            fill=white, fill opacity=0.75, text opacity=1,
            inner sep=1.5pt] at (img.north west) {\scriptsize (\alph{panel})};
    \end{tikzpicture}%
  \end{minipage}%
}

\begin{tabular}{c c c c c c c}
\rowlab{Conventional}{DFP (w/o silicone)} &
\labimg{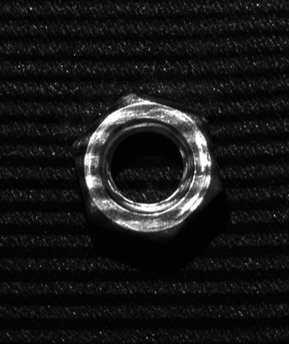}{nut} &
\labimg{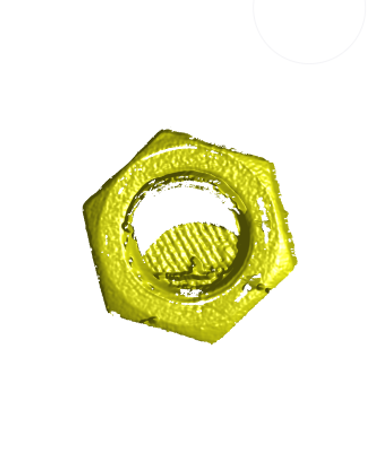}{nut3D} &
\labimg{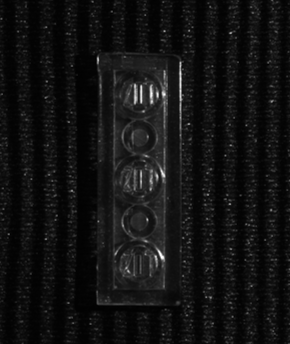}{lego} &
\labimg{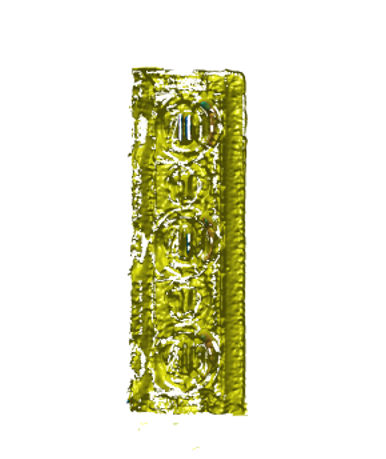}{lego3D} &
\labimg{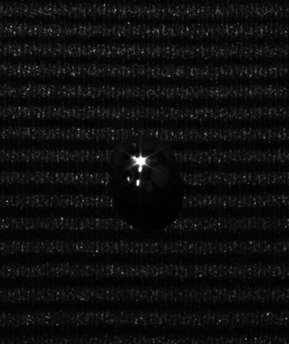}{ball} &
\labimg{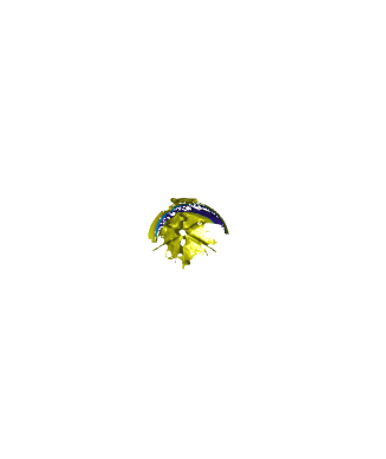}{ball3D} \\
\noalign{\vskip\rowgap}

\rowlab{Proposed}{method (contact-DFP)} &
\labimg{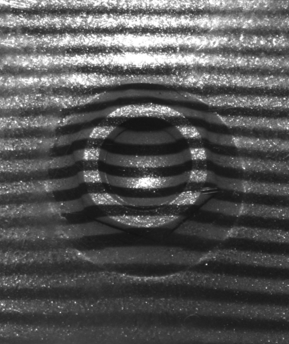}{nutsi} &
\labimg{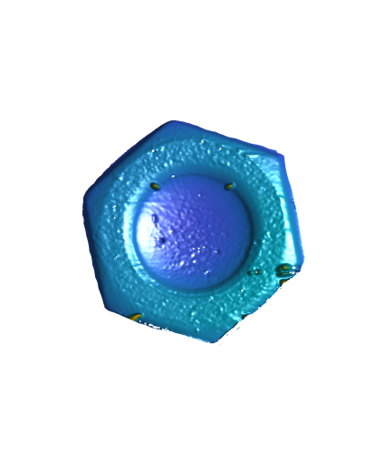}{nutsi3D} &
\labimg{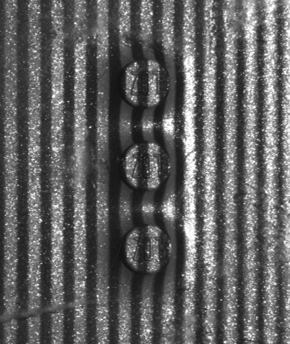}{legosi} &
\labimg{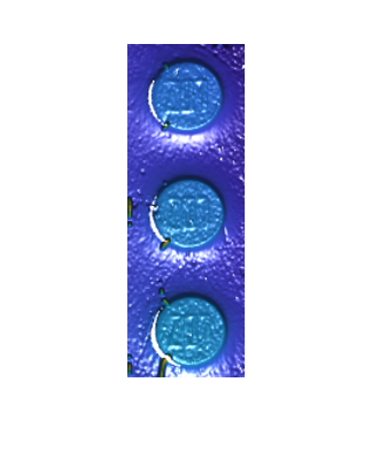}{legosi3D} &
\labimg{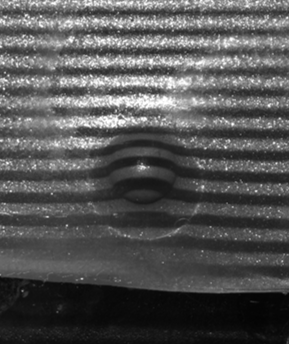}{ballsi} &
\labimg{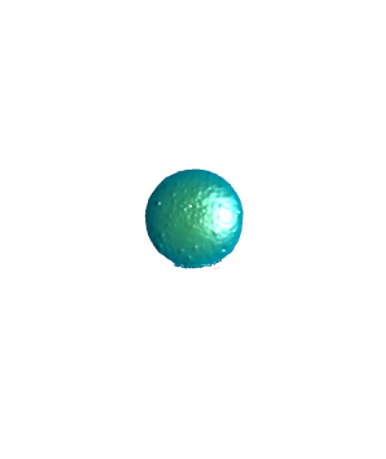}{ballsi3D} \\
\end{tabular}

\addtocounter{figure}{-1}
\caption{%
3D reconstruction comparison on three object types. For each object, the left panel shows the captured fringe image and the right panel shows the corresponding 3D reconstruction: nut (a,b,g,h), transparent LEGO block (c,d,i,j), and metallic ball (e,f,k,l). Top row: conventional DFP (direct projection, no silicone). Bottom row: proposed contact-DFP (coated silicone).}
\label{fig:fig5_subfig}

\end{figure*}
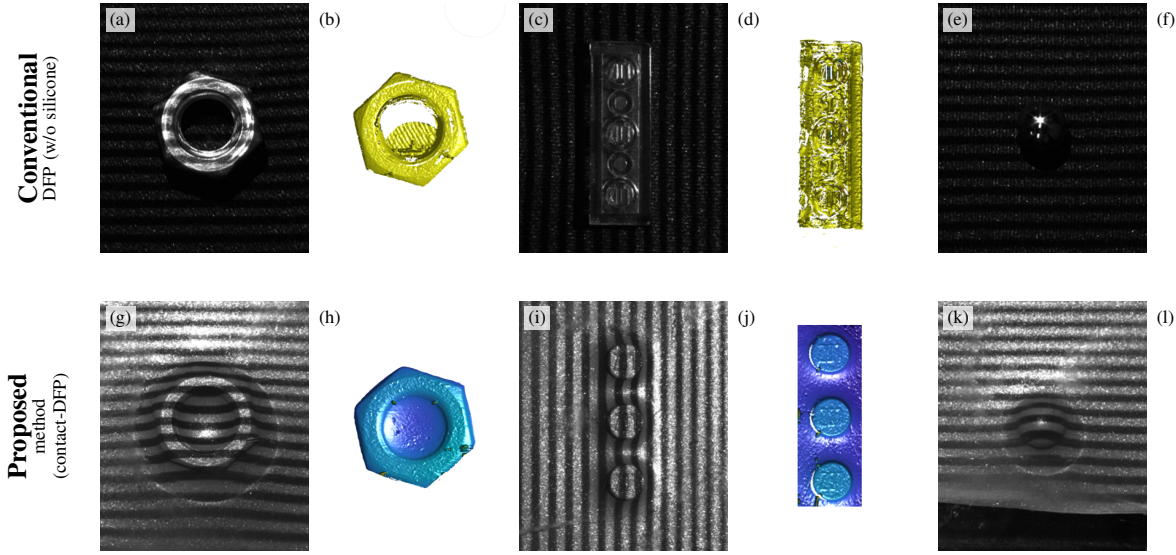

We conducted comparative experiments with images captured using a conventional DFP system. The objects used in the experiments are shown in Fig.~\ref{nut}, Fig.~\ref{lego}, and Fig.~\ref{ball}, including reflective metallic objects such as a nut and a metallic ball, as well as a transparent LEGO block. In the conventional DFP system, the phase-shifting method is used to obtain a phase map for 3-D reconstruction. However, as shown in Fig.~\ref{nut}, due to the strong reflection from the metallic nut, the projector light induces local saturation and prevents the fringe pattern from being fully formed on the object surface. Consequently, the reconstruction result shown in Fig.~\ref{nut3D} is incomplete.

To address this issue, we placed the object on the stage designed in our setup, as shown in Fig.~\ref{fig:Setup}, and elevated it to make contact with the fixed layer of fabricated silicone. Fig.~\ref{nutsi} shows the image captured during the experiment using the proposed method, where the fringe patterns are correctly projected onto the deformed coated silicone surface. The corresponding 3-D reconstruction result using the proposed method is shown in Fig.~\ref{nutsi3D}.

For the transparent object, we used a transparent LEGO block. Fig.~\ref{lego} shows the image captured using the conventional DFP system. Due to the transparency of the object, the projected fringe patterns are transmitted through the object and distorted by internal reflection, which prevents reliable phase retrieval and results in incomplete reconstruction, as shown in Fig.~\ref{lego3D}. In contrast, when the experiment was conducted using the proposed method, the fringe patterns were clearly formed on the coated silicone surface, as shown in Fig.~\ref{legosi}, and the corresponding 3-D reconstruction was successfully obtained, as shown in Fig.~\ref{legosi3D}.

Finally, Fig.~\ref{ball} shows the image of a metallic ball captured using the conventional DFP system. Because of its highly reflective and curved surface, the projected fringe patterns are distorted or saturated, leading to incomplete 3-D reconstruction, as shown in Fig.~\ref{ball3D}. However, when the metallic ball was tested using the proposed method, fringe patterns were successfully formed on the coated silicone surface, as shown in Fig.~\ref{ballsi}, and a stable 3-D reconstruction was obtained, as shown in Fig.~\ref{ballsi3D}.

In conclusion, the conventional DFP system fails to achieve proper 3-D reconstruction for reflective metallic objects and transparent objects because fringe patterns are not correctly formed on the target surfaces. The proposed method effectively resolves these issues by projecting fringe patterns onto the deformed coated silicone surface, enabling stable 3-D reconstruction of optically challenging objects.

\subsection{Dimensional evaluation using acrylic rods}

Conventional DFP methods often encounter difficulties in accurately reconstructing metallic and transparent objects because of light saturation, reflection, and projection distortion. In contrast, the proposed method successfully provided stable 3-D reconstructions for such optically challenging targets. To further examine the dimensional reliability of the proposed method using calibrated transparent artifacts, experiments were conducted using three acrylic rods with reference widths of 1.99, 4.81, and 7.23 mm, as measured by a vernier caliper.

\begin{figure}
\centering
\includegraphics[width=2.259in, height=1.5in]{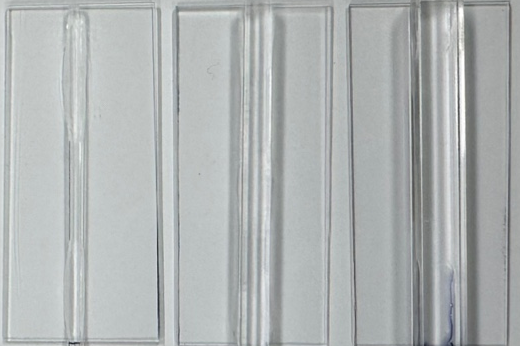}%
\caption{Three types of acrylic rods with different sizes}
\label{acrylic3}
\end{figure}

In this experiment, each acrylic rod was reconstructed over five repeated trials, and the reconstructed cross-sectional widths were compared with the corresponding reference dimensions. In addition to the dimensional comparison, the reconstructed cross-sectional profiles were also examined to verify whether the rod geometries were properly recovered. Figure~\ref{acrylic3} shows the three acrylic rods used in the experiment. The rods were cut into manageable pieces and fixed on an acrylic plate to ensure a stable measurement condition.

\begin{figure}
\centering
\subfloat[]{\includegraphics[width=0.85in, height=1.3in]{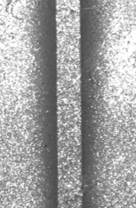}\label{2mm}}%
\hspace{0.005in}
\subfloat[]{\includegraphics[width=0.85in, height=1.3in]{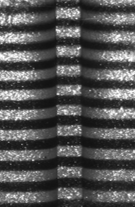}\label{2mmfringe}}%
\hspace{0.005in}
\subfloat[]{\includegraphics[width=0.85in, height=1.3in]{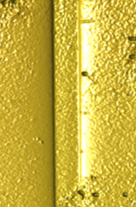}\label{2mm3D}}%

\subfloat[]{\includegraphics[width=3.2in, height=0.699in]{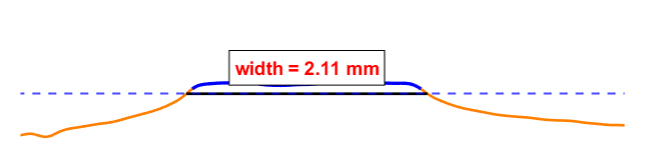}\label{2mm_reconstruction}}%

\subfloat[]{\includegraphics[width=0.85in, height=1.3in]{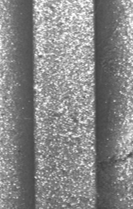}\label{5mm}}%
\hspace{0.005in}
\subfloat[]{\includegraphics[width=0.85in, height=1.3in]{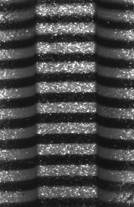}\label{5mmfringe}}%
\hspace{0.005in}
\subfloat[]{\includegraphics[width=0.85in, height=1.3in]{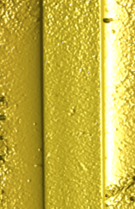}\label{5mm3D}}%

\subfloat[]{\includegraphics[width=3.2in, height=0.699in]{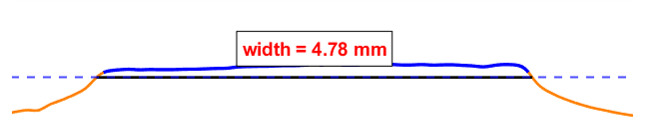}\label{5mm_reconstruction}}%

\subfloat[]{\includegraphics[width=0.85in, height=1.3in]{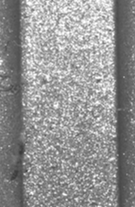}\label{8mm}}%
\hspace{0.005in}
\subfloat[]{\includegraphics[width=0.85in, height=1.3in]{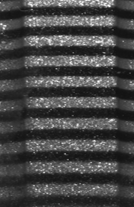}\label{8mmfringe}}%
\hspace{0.005in}
\subfloat[]{\includegraphics[width=0.85in, height=1.3in]{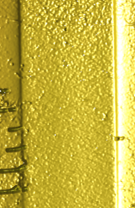}\label{8mm3D}}%

\subfloat[]{\includegraphics[width=3.2in, height=0.699in]{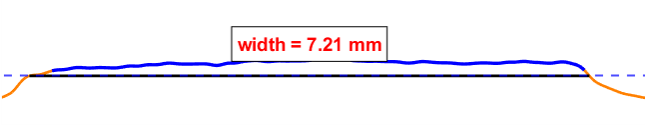}\label{8mm_reconstruction}}%

\caption{Dimensional evaluation of acrylic rods: (a), (e), and (i) captured images of the acrylic rods; (b), (f), and (j) corresponding fringe images; (c), (g), and (k) reconstructed 3-D surfaces; and (d), (h), and (l) cross-sectional profiles used to measure the rod thicknesses.}
\label{fig_Acrylic}
\end{figure}

Figure~\ref{2mm}--\ref{2mm3D} show a representative example from the repeated trials for the 1.99 mm acrylic rod, including the actual image, fringe image, and reconstructed 3-D result. The representative cross-sectional measurement shown in Fig.~\ref{2mm_reconstruction} was 2.11 mm. Likewise, Fig.~\ref{5mm}--\ref{5mm3D} present a representative example for the 4.81 mm acrylic rod, and the corresponding cross-sectional measurement shown in Fig.~\ref{5mm_reconstruction} was 4.78 mm. Similarly, Fig.~\ref{8mm}--\ref{8mm3D} show a representative example for the 7.23 mm acrylic rod, with a representative cross-sectional measurement of 7.21 mm in Fig.~\ref{8mm_reconstruction}. These figures are shown as representative examples only, while the overall dimensional results from all five repeated trials are summarized in Table~\ref{tab:rod_repeat}.

As summarized in Table~\ref{tab:rod_repeat}, the reconstructed widths showed reasonable agreement with the reference values across rods of different sizes, while the trial-to-trial variation remained limited. In addition, the reconstructed cross-sectional profiles exhibited well-defined rod boundaries, indicating that the proposed method can recover physically meaningful cross-sectional geometry for transparent contact objects. The close agreement with the caliper reference also indicates that, under the light contact load used here, the lateral deformation of the silicone is small enough that the recovered cross-sectional width approximates the rigid-object dimension; this contrasts with the highly curved sphere case in Section~\ref{sec:sphere_eval}, where contact deformation introduces a more pronounced offset in the absolute radius. This experiment is intended as a repeatability-based dimensional evaluation rather than a full uncertainty analysis.

\begin{figure}[t]
\scriptsize 
\setlength{\tabcolsep}{5pt}     
\renewcommand{\arraystretch}{1.05} 

\centering
\begin{tabular}{>{\centering\arraybackslash}m{1.5cm} c c c}
   & \textbf{T-bolt} & \textbf{Transparent bear} & \textbf{Two balls} \\[4pt]

   \makecell{\textbf{Actual}\\\textbf{object}} &
   \adjustbox{valign=c}{\includegraphics[width=0.72in,height=0.95in,keepaspectratio]{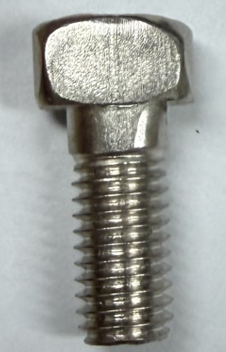}} &
   \adjustbox{valign=c}{\includegraphics[width=0.72in,height=0.95in,keepaspectratio]{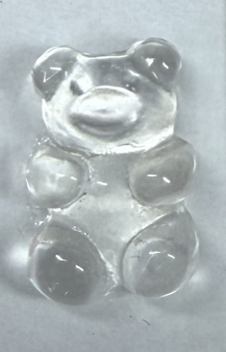}} &
   \adjustbox{valign=c}{\includegraphics[width=0.72in,height=0.95in,keepaspectratio]{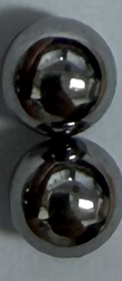}} \\[4pt]
   
   \makecell{\textbf{GelSight}\\\textbf{2-D}} &
   \adjustbox{valign=c}{\includegraphics[width=0.72in,height=0.95in,keepaspectratio]{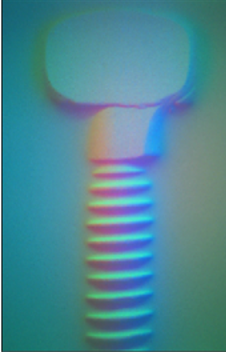}} &
   \adjustbox{valign=c}{\includegraphics[width=0.72in,height=0.95in,keepaspectratio]{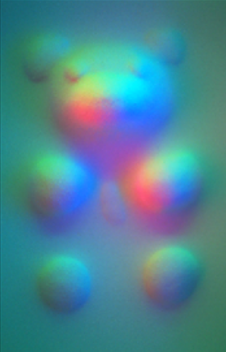}} &
   \adjustbox{valign=c}{\includegraphics[width=0.72in,height=0.95in,keepaspectratio]{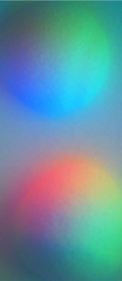}} \\[4pt]

   \makecell{\textbf{Fringe}\\\textbf{Image}} &
   \adjustbox{valign=c}{\includegraphics[width=0.72in,height=0.95in,keepaspectratio]{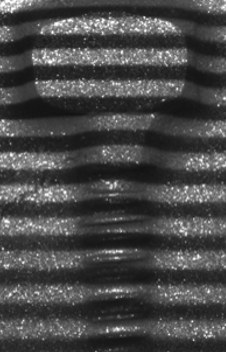}} &
   \adjustbox{valign=c}{\includegraphics[width=0.72in,height=0.95in,keepaspectratio]{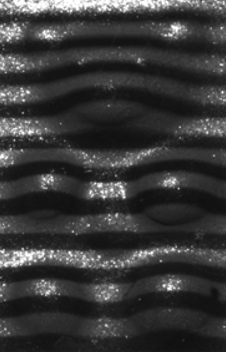}} &
   \adjustbox{valign=c}{\includegraphics[width=0.72in,height=0.95in,keepaspectratio]{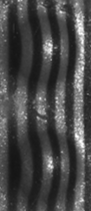}} \\[4pt]

    \makecell{\textbf{Gelsight}\\\textbf{3-D}} &
   \adjustbox{valign=c}{\includegraphics[width=0.72in,height=0.95in,keepaspectratio]{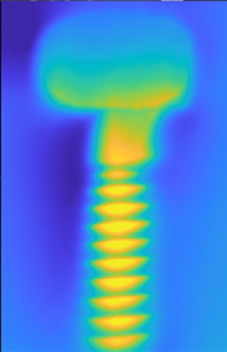}} &
   \adjustbox{valign=c}{\includegraphics[width=0.72in,height=0.95in,keepaspectratio]{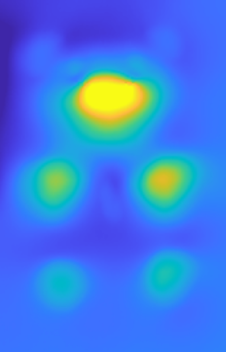}} &
   \adjustbox{valign=c}{\includegraphics[width=0.72in,height=0.95in,keepaspectratio]{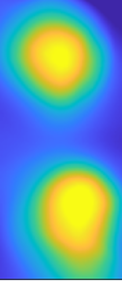}} \\

   \makecell{\textbf{Proposed}\\\textbf{3-D}} &
   \adjustbox{valign=c}{\includegraphics[width=0.72in,height=0.95in,keepaspectratio]{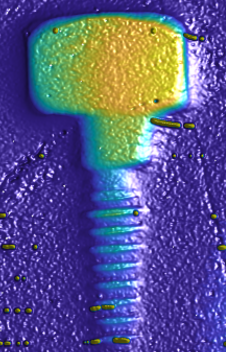}} &
   \adjustbox{valign=c}{\includegraphics[width=0.72in,height=0.95in,keepaspectratio]{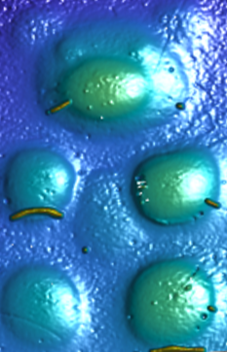}} &
   \adjustbox{valign=c}{\includegraphics[width=0.72in,height=0.95in,keepaspectratio]{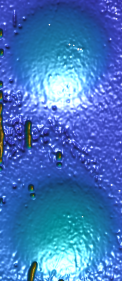}} \\

\end{tabular}

\caption{Grid comparison of methods vs. objects. Columns: T-bolt, transparent bear, and two balls. Rows (top to bottom): actual object, GelSight 2-D image, fringe image of the proposed method, GelSight 3-D reconstruction, and proposed 3-D reconstruction.}
\label{fig:comparison}
\end{figure}

\begin{table}[h]
\caption{Dimensional evaluation of reconstructed acrylic rods over five repeated trials.}
\label{tab:rod_repeat}
\centering
\footnotesize
\setlength{\tabcolsep}{4pt}
\renewcommand{\arraystretch}{1.15}
\begin{tabular}{c c c c c c}
\hline
Rod & Ref. (mm) & Mean (mm) & Abs. error (mm) & SD (mm) & CV (\%) \\
\hline
A & 1.99 & 1.970 & 0.020 & 0.085 & 4.31 \\
B & 4.81 & 4.756 & 0.054 & 0.081 & 1.70 \\
C & 7.23 & 7.072 & 0.158 & 0.224 & 3.17 \\
\hline
\end{tabular}
\vspace{2pt}
\begin{flushleft}\footnotesize
Ref. denotes vernier caliper measurements. SD is the sample standard deviation over five repeated trials, and CV denotes the coefficient of variation.
\end{flushleft}
\end{table}

\subsection{Comparative Analysis with Gelsight}

To further validate the performance of the proposed method, we conducted additional experiments using a GelSight Mini sensor and performed side-by-side comparisons. The test objects were a metallic T-bolt and a transparent bear, together with a connected two-sphere object composed of two joined 8 mm metallic spheres, as shown in Fig.~\ref{fig:comparison}. In particular, the two-sphere configuration was included to assess reconstruction robustness under strong local curvature variation and partial self-occlusion. As summarized in Fig.~\ref{fig:comparison}, GelSight recovers the overall appearance of each contact but tends to lose fine relief and to flatten regions of strong curvature, whereas the proposed contact-DFP reconstruction preserves the contact geometry more consistently across all three objects. This qualitative comparison is quantified in the sphere-fitting evaluation presented next.

\subsection{Sphere-Fitting-Based Quantitative Evaluation Against GelSight}
\label{sec:sphere_eval}

To quantitatively evaluate the reconstruction accuracy of the proposed tactile 3-D measurement method, a sphere-fitting-based comparison was performed against a GelSight Mini sensor. The test specimen consisted of two connected metallic spheres with a nominal diameter of 8 mm, corresponding to a nominal radius of 4 mm. This specimen was selected because the spherical geometry provides an analytic reference shape while also introducing strong local curvature and contact deformation, which are challenging conditions for vision-based tactile reconstruction.

It is important to clarify the measurand for this evaluation. As in other elastomer-based tactile sensors, the proposed method does not image the rigid object directly; it reconstructs the contact relief imprinted on the deformable silicone surface. Accordingly, the primary measurand is the fidelity with which this contact surface is recovered, rather than the absolute dimension of the underlying rigid sphere. Two complementary classes of metrics are therefore reported. The radial residual metrics (RMSE, MAE, and P95), which quantify how consistently the reconstructed points conform to a fitted sphere, are independent of any constant offset and thus directly measure reconstruction fidelity; they constitute the primary accuracy indicator. The fitted radius and absolute radius error, by contrast, are reported as secondary indicators, since they additionally carry the systematic offset introduced by the physical deformation of the silicone layer, which neither method explicitly compensates.

For the GelSight comparison, the RGB tactile images were reconstructed using a recalibrated photometric-stereo pipeline. First, a reference sphere with known geometry was used to construct a lookup table (LUT) that maps the measured RGB response to local surface gradients. In the calibration image, the contact region of the reference sphere was segmented, and the corresponding surface gradients were analytically calculated from the known spherical geometry. Each valid calibration pixel therefore provided a correspondence between the measured RGB intensity and the local gradient components. The RGB values were then discretized into bins, and the gradient values assigned to similar RGB responses were averaged to generate the empirical LUT. Missing or sparsely populated bins were filled using neighboring valid bins. The calibrated LUT was subsequently applied to the RGB image of the test object to estimate the gradient field, which was integrated to obtain the GelSight height map.

For both the GelSight reconstruction and the proposed reconstruction, the two spherical contact regions were processed using the same evaluation procedure. Each reconstructed surface was first converted into a metric point cloud. For the proposed method, the 3-D coordinates were directly obtained from the calibrated camera--projector triangulation. For GelSight, the lateral pixel coordinates and the reconstructed height map were converted into millimeter units based on the calibration result. The valid points corresponding to each sphere were then extracted, and least-squares sphere fitting was independently applied to each point cloud. The fitted radius, fitted diameter, radius error, root-mean-square error (RMSE), mean absolute error (MAE), and 95th-percentile absolute radial residual (P95) were calculated from the radial residuals between the reconstructed points and the fitted sphere.

\begin{figure*}[t]
\centering

\subfloat[GelSight reconstructed height surface.]{
\includegraphics[width=3.35in,height=1.65in,keepaspectratio]{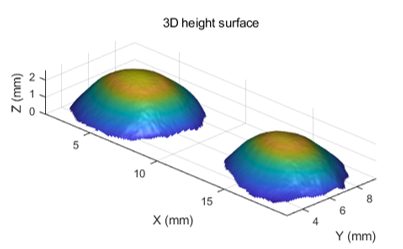}
\label{fig:sphere_gelsight_surface}
}
\hspace{0.12in}
\subfloat[Proposed reconstructed height surface.]{
\includegraphics[width=3.35in,height=1.65in,keepaspectratio]{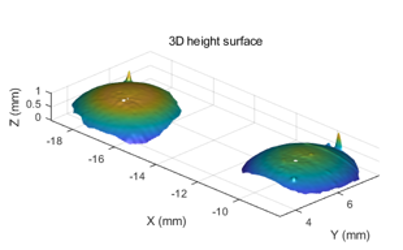}
\label{fig:sphere_proposed_surface}
}

\vspace{2mm}

\subfloat[GelSight sphere-fitting residual maps.]{
\includegraphics[width=3.35in,height=1.45in,keepaspectratio]{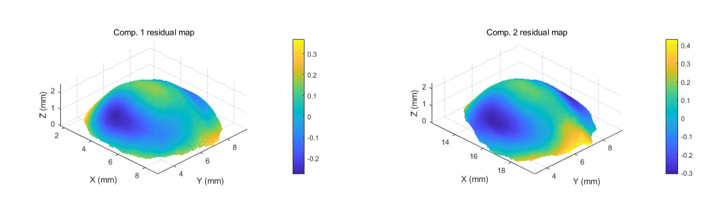}
\label{fig:sphere_gelsight_residual_map}
}
\hfill
\subfloat[Proposed sphere-fitting residual maps.]{
\includegraphics[width=3.35in,height=1.45in,keepaspectratio]{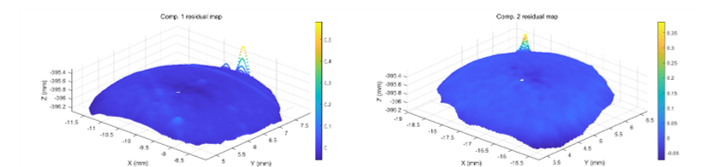}
\label{fig:sphere_proposed_residual_map}
}

\vspace{2mm}

\subfloat[GelSight radial residual histograms.]{
\includegraphics[width=3.35in,height=1.35in,keepaspectratio]{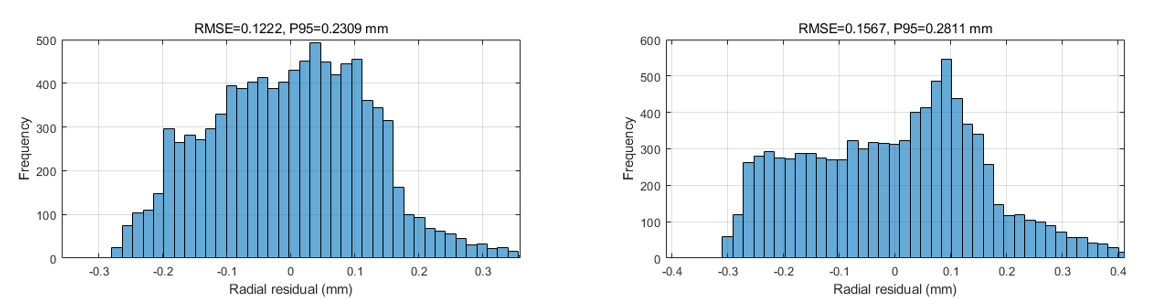}
\label{fig:sphere_gelsight_histogram}
}
\hfill
\subfloat[Proposed radial residual histograms.]{
\includegraphics[width=3.35in,height=1.35in,keepaspectratio]{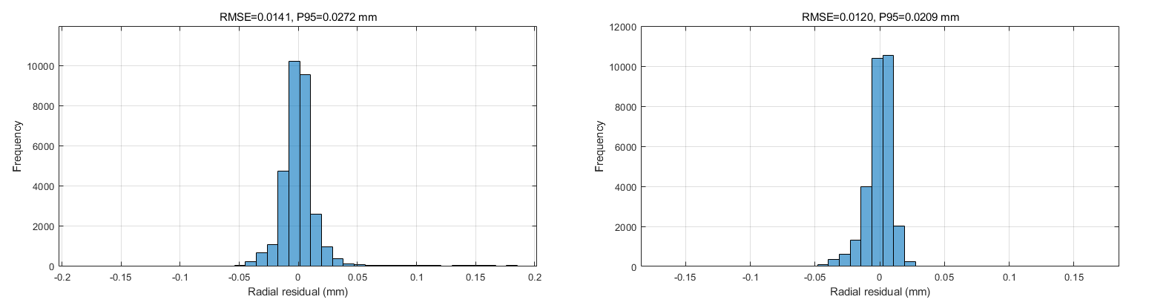}
\label{fig:sphere_proposed_histogram}
}

\caption{Sphere-fitting-based quantitative comparison between GelSight and the proposed method using two connected 8 mm metallic spheres. 
(a) and (b) show the reconstructed height surfaces obtained from GelSight and the proposed method, respectively. 
(c) and (d) show the corresponding radial residual maps after least-squares sphere fitting for the two spherical components. 
(e) and (f) show the radial residual histograms. 
Compared with GelSight, the proposed method exhibits more compact residual distributions and smaller spatial variation in the residual maps, indicating more stable spherical reconstruction under the same silicone-contact condition.}
\label{fig:sphere_fitting_comparison}
\end{figure*}

Fig.~\ref{fig:sphere_fitting_comparison} shows the sphere-fitting results for GelSight and the proposed method. The GelSight reconstruction produced the overall spherical shape, but the reconstructed surfaces exhibited relatively large radial deviations, especially near regions with strong curvature and contact-induced intensity variation. This behavior is also observed in the residual maps and histograms, where the radial residuals are broadly distributed. In contrast, the proposed method produced more compact residual distributions for both spheres. The residual maps show smaller spatial variation, and the residual histograms are concentrated near zero, indicating that the reconstructed surface is more consistent with a spherical geometry.

\begin{table*}[t]
\centering
\caption{Sphere-fitting comparison between GelSight reconstruction and the proposed method using two connected 8 mm metallic spheres. The nominal sphere radius is 4 mm. Lower values are better for absolute radius error, RMSE, MAE, and P95.}
\label{tab:sphere_fitting_comparison}
\resizebox{\textwidth}{!}{%
\begin{tabular}{llcccccc}
\toprule
Method & Sphere 
& Fitted radius (mm) 
& Fitted diameter (mm) 
& Abs. radius error (mm) 
& RMSE (mm) 
& MAE (mm) 
& P95 (mm) \\
\midrule
GelSight & Sphere 1 & 3.609 & 7.219 & 0.391 & 0.1222 & 0.1008 & 0.2309 \\
GelSight & Sphere 2 & 3.682 & 7.365 & 0.318 & 0.1567 & 0.1309 & 0.2811 \\
GelSight & Mean     & 3.646 & 7.292 & 0.354 & 0.1394 & 0.1159 & 0.2560 \\
\midrule
Proposed & Sphere 1 & 3.758 & 7.516 & 0.242 & 0.0141 & 0.0091 & 0.0272 \\
Proposed & Sphere 2 & 3.910 & 7.820 & 0.090 & 0.0120 & 0.0075 & 0.0209 \\
Proposed & Mean     & 3.834 & 7.668 & 0.166 & 0.0130 & 0.0083 & 0.0241 \\
\bottomrule
\end{tabular}%
}
\end{table*}

The quantitative results are summarized in Table~\ref{tab:sphere_fitting_comparison}. For GelSight, the fitted radii were 3.609 mm and 3.682 mm for Spheres 1 and 2, respectively, resulting in a mean fitted radius of 3.646 mm and a mean absolute radius error of 0.354 mm. The corresponding mean RMSE, MAE, and P95 values were 0.1394 mm, 0.1159 mm, and 0.2560 mm, respectively. In comparison, the proposed method yielded fitted radii of 3.758 mm and 3.910 mm, with a mean fitted radius of 3.834 mm and a mean absolute radius error of 0.166 mm. The mean RMSE, MAE, and P95 values were reduced to 0.0130 mm, 0.0083 mm, and 0.0241 mm, respectively.

These results indicate that the proposed method provides a more stable spherical reconstruction than GelSight under the same silicone-contact condition. The primary fidelity indicator, the radial residual, improved markedly: the proposed method reduced the mean sphere-fitting RMSE by approximately one order of magnitude compared with GelSight, with consistent improvements in the MAE and P95 metrics. Because these residual metrics are invariant to a constant radial offset, this gain reflects a genuine improvement in reconstruction fidelity and is not affected by the silicone-deformation bias discussed above. The fitted radius, reported as a secondary indicator, should be interpreted as the apparent spherical curvature of the deformed silicone surface rather than the exact rigid-sphere radius, since neither method explicitly compensates for the physical deformation of the silicone layer. Even so, under the same contact and evaluation conditions, the proposed method also produced fitted radii closer to the nominal sphere geometry than GelSight.

\subsection{Uncertainty Analysis of Sphere-Fitting Measurements}

To further assess the metrological reliability of the sphere-fitting results, an uncertainty analysis was additionally performed based on repeated measurements of the sphere-fitting experiment. In this analysis, the measurand was defined as the fitted sphere radius obtained from the reconstructed 3-D surface of each sphere. Repeated trials were conducted by re-acquiring the contact data under the same nominal experimental condition, and the entire processing pipeline, including reconstruction, preprocessing, region-of-interest selection, and least-squares sphere fitting, was repeated for each trial. The radius error with respect to the reference sphere radius was also evaluated together with the fitted radius.

The combined standard uncertainty was estimated by considering three main contributions: repeatability of repeated trials, reference uncertainty of the sphere artifact, and calibration-related uncertainty associated with metric scaling. This uncertainty-budget formulation follows the general approach adopted in optical and structured-light-based metrology systems~\cite{huang2023three,zheng2022universal}. Accordingly, the combined standard uncertainty was calculated as
\begin{equation}
u_c =
\sqrt{
u_{\mathrm{rep}}^2
+
u_{\mathrm{ref}}^2
+
u_{\mathrm{cal}}^2
},
\end{equation}
where $u_{\mathrm{rep}}$ denotes the standard uncertainty due to measurement repeatability, $u_{\mathrm{ref}}$ denotes the uncertainty of the reference sphere dimension, and $u_{\mathrm{cal}}$ denotes the uncertainty associated with metric calibration or scale conversion.

The standard uncertainty associated with the reference sphere was estimated based on the nominal tolerance of the sphere diameter. As the exact specification of the reference sphere was not available, a conservative tolerance of $\pm 0.01$ mm was assumed, which is consistent with typical specifications of commercially available precision spheres. Assuming a uniform distribution, the standard uncertainty was calculated as
\begin{equation}
u_{\mathrm{ref}} =
\frac{0.01}{\sqrt{3}}
=
0.0058~\mathrm{mm}.
\end{equation}
This value is significantly smaller than the repeatability uncertainty and therefore does not dominate the combined uncertainty.

The repeatability component was obtained from the repeated sphere-fitting results. If $R_i$ denotes the fitted radius from the $i$th repeated trial and $n$ is the number of trials, the sample mean $\bar{R}$ and the repeatability-based standard uncertainty were calculated as
\begin{equation}
\bar{R}
=
\frac{1}{n}
\sum_{i=1}^{n} R_i,
\end{equation}
\begin{equation}
u_{\mathrm{rep}}
=
\frac{s(R)}{\sqrt{n}},
\quad
s(R)
=
\sqrt{
\frac{1}{n-1}
\sum_{i=1}^{n}
\left(R_i-\bar{R}\right)^2
}.
\end{equation}
The radius error relative to the reference value $R_{\mathrm{ref}}$ was defined as
\begin{equation}
e_R = \bar{R} - R_{\mathrm{ref}}.
\end{equation}

The standard uncertainty associated with calibration was estimated from the metric calibration or scale-conversion process required to express the reconstructed surface in millimeter units. For the proposed method, this term is related to the camera--projector calibration and triangulation-based metric reconstruction. For the GelSight reconstruction, it is related to the RGB-to-gradient calibration and the metric scale conversion of the reconstructed height map. In this study, the resulting calibration-related standard uncertainty was set to
\begin{equation}
u_{\mathrm{cal}} = 0.003~\mathrm{mm}.
\end{equation}

The expanded uncertainty was then obtained using a coverage factor of $k=2$ as
\begin{equation}
U = k u_c, \quad k=2,
\end{equation}
which approximately corresponds to a confidence level of 95\% under the normality assumption.

Table~\ref{tab:repeated_sphere_fitting} summarizes the repeated sphere-fitting results obtained from five repeated trials for the connected two-sphere specimen. The proposed method yielded mean fitted radii of 3.835 mm and 3.846 mm for Spheres 1 and 2, respectively, whereas GelSight yielded 3.671 mm and 3.656 mm. The corresponding radius errors relative to the reference radius of 4 mm were -0.165 mm and -0.154 mm for the proposed method, while GelSight showed larger errors of -0.329 mm and -0.344 mm. These results indicate that the proposed method produced fitted radii closer to the nominal sphere geometry under repeated contact conditions.

The uncertainty budget is summarized in Table~\ref{tab:uncertainty_budget}. The repeatability-based standard uncertainty was calculated from the sample standard deviation of the repeated fitted radii. For GelSight, the repeatability-based standard uncertainties were 0.030 mm and 0.040 mm for Spheres 1 and 2, respectively. For the proposed method, the corresponding values were reduced to 0.023 mm and 0.029 mm. In both methods, the reference and calibration-related uncertainty components were smaller than the repeatability component, indicating that the repeated contact and reconstruction process was the dominant contributor to the combined uncertainty.

\begin{table}[t]
\caption{Repeated sphere-fitting results for the connected two-sphere specimen.}
\label{tab:repeated_sphere_fitting}
\centering
\footnotesize
\setlength{\tabcolsep}{3pt}
\renewcommand{\arraystretch}{1.1}
\resizebox{\columnwidth}{!}{%
\begin{tabular}{c c c c c c}
\hline
Method & Sphere & $n$ & Mean radius & Error & SD \\
 &  &  & (mm) & (mm) & (mm) \\
\hline
GelSight & 1 & 5 & 3.671 & -0.329 & 0.067 \\
GelSight & 2 & 5 & 3.656 & -0.344 & 0.090 \\
Proposed & 1 & 5 & 3.835 & -0.165 & 0.052 \\
Proposed & 2 & 5 & 3.846 & -0.154 & 0.064 \\
\hline
\end{tabular}%
}
\end{table}

\begin{table}[t]
\caption{Uncertainty budget of the sphere-fitting measurements.}
\label{tab:uncertainty_budget}
\centering
\footnotesize
\setlength{\tabcolsep}{3pt}
\renewcommand{\arraystretch}{1.1}
\resizebox{\columnwidth}{!}{%
\begin{tabular}{c c c c c c}
\hline
Method & Sphere & $u_{\mathrm{rep}}$ & $u_{\mathrm{ref}}$ & $u_{\mathrm{cal}}$ & $U(k=2)$ \\
 &  & (mm) & (mm) & (mm) & (mm) \\
\hline
GelSight & 1 & 0.030 & 0.0058 & 0.003 & 0.061 \\
GelSight & 2 & 0.040 & 0.0058 & 0.003 & 0.082 \\
Proposed & 1 & 0.023 & 0.0058 & 0.003 & 0.048 \\
Proposed & 2 & 0.029 & 0.0058 & 0.003 & 0.059 \\
\hline
\end{tabular}%
}
\end{table}

Here, $u_{\mathrm{rep}}$ is the repeatability-based standard uncertainty obtained from repeated measurements, $u_{\mathrm{ref}}$ is the uncertainty of the reference sphere dimension, and $u_{\mathrm{cal}}$ is the calibration-related standard uncertainty.

The resulting expanded uncertainties were 0.048 mm and 0.059 mm for the proposed method for Spheres 1 and 2, respectively, whereas GelSight showed expanded uncertainties of 0.061 mm and 0.082 mm. Therefore, the proposed method showed both lower radius bias and lower expanded uncertainty than GelSight in the repeated sphere-fitting evaluation. These results demonstrate that the proposed triangulation-based tactile measurement provides more stable sphere-radius estimation under repeated contact and reconstruction conditions.

Overall, the uncertainty analysis supports that the proposed method provides reliable repeated sphere-fitting measurements for highly curved contact surfaces. In particular, the expanded uncertainty of the proposed method remained below 0.060 mm for both spheres, while the fitted radius was also closer to the nominal radius than that obtained from GelSight. These results indicate that the proposed contact-DFP framework improves not only the sphere-fitting residual performance but also the metrological stability of repeated tactile 3-D reconstruction.

\section{Conclusion}
In this paper, we proposed a tactile 3-D surface measurement method based on a DFP system with a coated silicone interface. The proposed approach enables stable reconstruction of surface geometry for optically challenging objects, including reflective and transparent materials, where conventional optical methods often fail.

Experimental results demonstrated that the proposed method significantly improves reconstruction completeness and reduces systematic bias compared with a vision-based tactile sensor (GelSight), while showing lower measurement variability across repeated trials.

Repeated sphere-fitting experiments further confirmed that the proposed method provides stable radius estimation, with lower measurement variability and expanded uncertainties below 0.060 mm for both spheres. These results confirm that the proposed method provides not only improved reconstruction completeness but also quantitatively reliable geometric measurement capability for highly curved contact surfaces.

Several limitations remain and motivate future work. First, the reconstruction recovers the contact relief on the deformable silicone rather than the rigid object itself, so the absolute dimension of the underlying object carries a systematic offset that depends on the silicone deformation under load; a contact-mechanics-based deformation compensation model is needed to convert the apparent contact geometry into the true rigid-object dimension. Second, the quantitative evaluation was based on a limited number of repeated trials and a small set of calibrated artifacts; a larger number of trials and traceable reference standards would strengthen the statistical and metrological confidence. Third, the present study focuses on static contact measurement, and extending the framework toward dynamic or continuous-scanning operation remains an open direction.

Overall, the proposed contact-based DFP framework offers a robust and practical solution for high-accuracy 3-D surface measurement in scenarios where conventional optical approaches are limited.

\section*{Acknowledgements}
The schematic in Fig.~\ref{fig:concept} was created with the assistance of a generative artificial intelligence image-generation tool, for illustrative purposes only, to aid reader understanding of the sensing principle. The authors reviewed and verified the technical content of the figure and take full responsibility for it.

This research was supported by the Technology Innovation Program (Project Name: Development of AI autonomous continuous production system technology for gas turbine blade maintenance and regeneration for power generation, Project Number: RS-2025-25447257, Contribution Rate: 30\%) funded by the Ministry of Trade, Industry and Resources (MOTIR, Korea), the Culture, Sports and Tourism R\&D Program through the Korea Creative Content Agency grant funded by the Ministry of Culture, Sports and Tourism in 2024 (Project Name: Global Talent for Generative AI Copyright Infringement and Copyright Theft, Project Number: RS-2024-00398413, Contribution Rate: 50\%), and the National Research Foundation of Korea (NRF) grant funded by the Korea government (MSIT) (Project Number: RS-2026-25498577, Contribution Rate: 20\%).

\bibliography{ref}

\begin{thebibliography}{10}
\providecommand{\url}[1]{#1}
\csname url@samestyle\endcsname
\providecommand{\newblock}{\relax}
\providecommand{\bibinfo}[2]{#2}
\providecommand{\BIBentrySTDinterwordspacing}{\spaceskip=0pt\relax}
\providecommand{\BIBentryALTinterwordstretchfactor}{4}
\providecommand{\BIBentryALTinterwordspacing}{\spaceskip=\fontdimen2\font plus
\BIBentryALTinterwordstretchfactor\fontdimen3\font minus
  \fontdimen4\font\relax}
\providecommand{\BIBforeignlanguage}[2]{{%
\expandafter\ifx\csname l@#1\endcsname\relax
\typeout{** WARNING: IEEEtran.bst: No hyphenation pattern has been}%
\typeout{** loaded for the language `#1'. Using the pattern for}%
\typeout{** the default language instead.}%
\else
\language=\csname l@#1\endcsname
\fi
#2}}
\providecommand{\BIBdecl}{\relax}
\BIBdecl

\bibitem{li2017high}
B.~Li, Y.~An, D.~Cappelleri, J.~Xu, and S.~Zhang, ``High-accuracy, high-speed
  3d structured light imaging techniques and potential applications to
  intelligent robotics,'' \emph{International journal of intelligent robotics
  and applications}, vol.~1, no.~1, pp. 86--103, 2017.

\bibitem{yuan2017gelsight}
W.~Yuan, S.~Dong, and E.~H. Adelson, ``Gelsight: High-resolution robot tactile
  sensors for estimating geometry and force,'' \emph{Sensors}, vol.~17, no.~12,
  p. 2762, 2017.

\bibitem{ward2018tactip}
B.~Ward-Cherrier, N.~Pestell, L.~Cramphorn, B.~Winstone, M.~E. Giannaccini,
  J.~Rossiter, and N.~F. Lepora, ``The tactip family: Soft optical tactile
  sensors with 3d-printed biomimetic morphologies,'' \emph{Soft robotics},
  vol.~5, no.~2, pp. 216--227, 2018.

\bibitem{lambeta2020digit}
M.~Lambeta, P.-W. Chou, S.~Tian, B.~Yang, B.~Maloon, V.~R. Most, D.~Stroud,
  R.~Santos, A.~Byagowi, G.~Kammerer \emph{et~al.}, ``Digit: A novel design for
  a low-cost compact high-resolution tactile sensor with application to in-hand
  manipulation,'' \emph{IEEE Robotics and Automation Letters}, vol.~5, no.~3,
  pp. 3838--3845, 2020.

\bibitem{donlon2018gelslim}
E.~Donlon, S.~Dong, M.~Liu, J.~Li, E.~Adelson, and A.~Rodriguez, ``Gelslim: A
  high-resolution, compact, robust, and calibrated tactile-sensing finger,'' in
  \emph{2018 IEEE/RSJ International Conference on Intelligent Robots and
  Systems (IROS)}.\hskip 1em plus 0.5em minus 0.4em\relax IEEE, 2018, pp.
  1927--1934.

\bibitem{wettels2008biomimetic}
N.~Wettels, V.~J. Santos, R.~S. Johansson, and G.~E. Loeb, ``Biomimetic tactile
  sensor array,'' \emph{Advanced robotics}, vol.~22, no.~8, pp. 829--849, 2008.

\bibitem{schmitz2011methods}
A.~Schmitz, P.~Maiolino, M.~Maggiali, L.~Natale, G.~Cannata, and G.~Metta,
  ``Methods and technologies for the implementation of large-scale robot
  tactile sensors,'' \emph{IEEE Transactions on Robotics}, vol.~27, no.~3, pp.
  389--400, 2011.

\bibitem{zhang2006high}
S.~Zhang and P.~S. Huang, ``High-resolution, real-time three-dimensional shape
  measurement,'' \emph{Optical Engineering}, vol.~45, no.~12, pp.
  123\,601--123\,601, 2006.

\bibitem{zuo2018phase}
C.~Zuo, S.~Feng, L.~Huang, T.~Tao, W.~Yin, and Q.~Chen, ``Phase shifting
  algorithms for fringe projection profilometry: A review,'' \emph{Optics and
  lasers in engineering}, vol. 109, pp. 23--59, 2018.

\bibitem{geng2011structured}
J.~Geng, ``Structured-light 3d surface imaging: a tutorial,'' \emph{Advances in
  optics and photonics}, vol.~3, no.~2, pp. 128--160, 2011.

\bibitem{zhang2009high}
S.~Zhang and S.-T. Yau, ``High dynamic range scanning technique,''
  \emph{Optical Engineering}, vol.~48, no.~3, pp. 033\,604--033\,604, 2009.

\bibitem{feng2014general}
S.~Feng, Y.~Zhang, Q.~Chen, C.~Zuo, R.~Li, and G.~Shen, ``General solution for
  high dynamic range three-dimensional shape measurement using the fringe
  projection technique,'' \emph{Optics and Lasers in Engineering}, vol.~59, pp.
  56--71, 2014.

\bibitem{feng2018high}
S.~Feng, L.~Zhang, C.~Zuo, T.~Tao, Q.~Chen, and G.~Gu, ``High dynamic range 3d
  measurements with fringe projection profilometry: a review,''
  \emph{Measurement Science and Technology}, vol.~29, no.~12, p. 122001, 2018.

\bibitem{liu20113d}
G.-h. Liu, X.-Y. Liu, and Q.-Y. Feng, ``3d shape measurement of objects with
  high dynamic range of surface reflectivity,'' \emph{Applied optics}, vol.~50,
  no.~23, pp. 4557--4565, 2011.

\bibitem{guo2004gamma}
H.~Guo, H.~He, and M.~Chen, ``Gamma correction for digital fringe projection
  profilometry,'' \emph{Applied optics}, vol.~43, no.~14, pp. 2906--2914, 2004.

\bibitem{lei2015multi}
Z.~Lei, C.~Wang, and C.~Zhou, ``Multi-frequency inverse-phase fringe projection
  profilometry for nonlinear phase error compensation,'' \emph{Optics and
  Lasers in Engineering}, vol.~66, pp. 249--257, 2015.

\bibitem{li2022exposure}
J.~Li, J.~Guan, X.~Chen, X.~Le, and J.~Xi, ``Exposure map fusion for precise
  3-d reconstruction of high dynamic range surfaces,'' \emph{IEEE Transactions
  on Instrumentation and Measurement}, vol.~71, pp. 1--11, 2022.

\bibitem{feng2024multi}
Y.~Feng, R.~Wu, P.~Li, W.~Wu, J.~Lin, X.~Liu, and L.~Chen, ``Multi-view
  high-dynamic-range 3d reconstruction and point cloud quality evaluation based
  on dual-frame difference images,'' \emph{Applied Optics}, vol.~63, no.~30,
  pp. 7865--7874, 2024.

\bibitem{wang2023saturation}
J.~Wang, Y.~Yang, P.~Xu, and J.~Liu, ``Saturation-induced phase error
  correction method in 3-d measurement based on inverted fringes,''
  \emph{Applied Optics}, vol.~62, no.~2, pp. 492--499, 2023.

\bibitem{wei2024method}
H.~Wei, H.~Li, J.~Liu, G.~Deng, and S.~Zhou, ``A method for suppressing
  saturation-induced errors in fringe projection profilometry by fringe
  restoration assisted via euler formula-based method and intersection
  points,'' \emph{IEEE Transactions on Instrumentation and Measurement}, 2024.

\bibitem{li2023fringe}
H.~Li, H.~Wei, J.~Liu, G.~Deng, S.~Zhou, W.~Wang, L.~He, and P.~Tian, ``Fringe
  projection profilometry based on saturated fringe restoration in high dynamic
  range scenes,'' \emph{Sensors}, vol.~23, no.~6, p. 3133, 2023.

\bibitem{woodham1980photometric}
R.~J. Woodham, ``Photometric method for determining surface orientation from
  multiple images,'' \emph{Optical engineering}, vol.~19, no.~1, pp. 139--144,
  1980.

\bibitem{mannsfeld2010highly}
S.~C. Mannsfeld, B.~C. Tee, R.~M. Stoltenberg, C.~V.~H. Chen, S.~Barman, B.~V.
  Muir, A.~N. Sokolov, C.~Reese, and Z.~Bao, ``Highly sensitive flexible
  pressure sensors with microstructured rubber dielectric layers,''
  \emph{Nature materials}, vol.~9, no.~10, pp. 859--864, 2010.

\bibitem{wang2013user}
C.~Wang, D.~Hwang, Z.~Yu, K.~Takei, J.~Park, T.~Chen, B.~Ma, and A.~Javey,
  ``User-interactive electronic skin for instantaneous pressure
  visualization,'' \emph{Nature materials}, vol.~12, no.~10, pp. 899--904,
  2013.

\bibitem{park2016mos2}
M.~Park, Y.~J. Park, X.~Chen, Y.-K. Park, M.-S. Kim, and J.-H. Ahn,
  ``Mos2-based tactile sensor for electronic skin applications,''
  \emph{Advanced Materials}, vol.~28, no.~13, pp. 2556--2562, 2016.

\bibitem{jiang2024electrical}
Z.~Jiang, Z.~Xu, M.~Li, H.~Zeng, F.~Gong, and Y.~Tang, ``Electrical impedance
  tomography-based electronic skin for multi-touch tactile sensing using
  hydrogel material and fista algorithm,'' \emph{Sensors}, vol.~24, no.~18, p.
  5985, 2024.

\bibitem{andrussow2023minsight}
I.~Andrussow, H.~Sun, K.~J. Kuchenbecker, and G.~Martius, ``Minsight: A
  fingertip-sized vision-based tactile sensor for robotic manipulation,''
  \emph{Advanced Intelligent Systems}, vol.~5, no.~8, p. 2300042, 2023.

\bibitem{kim2024extremely}
K.~Kim, J.-H. Hong, K.~Bae, K.~Lee, D.~J. Lee, J.~Park, H.~Zhang, M.~Sang,
  J.~E. Ju, Y.~U. Cho \emph{et~al.}, ``Extremely durable electrical impedance
  tomography--based soft and ultrathin wearable e-skin for three-dimensional
  tactile interfaces,'' \emph{Science Advances}, vol.~10, no.~38, p. eadr1099,
  2024.

\bibitem{yeo2025high}
I.~Yeo and J.-S. Hyun, ``High-resolution tactile sensor for 3{D} surface
  measurement using structured light system,'' in \emph{Dimensional Optical
  Metrology and Inspection for Practical Applications XIV}, vol. 13462.\hskip
  1em plus 0.5em minus 0.4em\relax SPIE, 2025, p. 134620F.

\bibitem{zhang2000flexible}
Z.~Zhang, ``A flexible new technique for camera calibration,'' \emph{IEEE
  Transactions on pattern analysis and machine intelligence}, vol.~22, no.~11,
  pp. 1330--1334, 2000.

\bibitem{abad2019pilot}
A.~C. Abad, D.~Reid, and A.~Ranasinghe, ``Pilot study: Low cost gelsight
  sensor,'' in \emph{Workshop on ViTac: Integrating Vision and Touch for
  Multimodal and Cross-modal Perception, ICRA}, 2019.

\bibitem{zhao2023gelsight}
J.~Zhao and E.~H. Adelson, ``Gelsight svelte: A human finger-shaped
  single-camera tactile robot finger with large sensing coverage and
  proprioceptive sensing,'' in \emph{2023 IEEE/RSJ International Conference on
  Intelligent Robots and Systems (IROS)}.\hskip 1em plus 0.5em minus
  0.4em\relax IEEE, 2023, pp. 8979--8984.

\bibitem{zhang2018highbook}
S.~Zhang, \emph{High-speed 3D imaging with digital fringe projection
  techniques}.\hskip 1em plus 0.5em minus 0.4em\relax CRC Press, 2018.

\bibitem{zuo2016temporal}
C.~Zuo, L.~Huang, M.~Zhang, Q.~Chen, and A.~Asundi, ``Temporal phase unwrapping
  algorithms for fringe projection profilometry: A comparative review,''
  \emph{Optics and lasers in engineering}, vol.~85, pp. 84--103, 2016.

\bibitem{huang2023three}
J.~Huang, R.~Yang, D.~Lian, J.~Liu, and J.~Tan, ``Three-dimensional method
  combining linearly structured light sensing and rotary scanning for measuring
  aviation bearings,'' \emph{IEEE Transactions on Instrumentation and
  Measurement}, vol.~72, pp. 1--10, 2023.

\bibitem{zheng2022universal}
Y.~Zheng, M.~Duan, Z.~Sun, X.~Fan, Y.~Jin, J.~Zheng, C.~Zhu, and E.~Chen, ``A
  universal self-correcting approach for abnormal jump errors in absolute phase
  retrieval,'' \emph{IEEE Transactions on Instrumentation and Measurement},
  vol.~71, pp. 1--13, 2022.

\end{thebibliography}
\bibliographystyle{IEEEtran}

\end{document}